\documentclass[a4paper]{article}

\usepackage{INTERSPEECH2022}
\usepackage{times}
\usepackage{booktabs}
\usepackage{latexsym}
\usepackage{graphicx}
\usepackage{tabularx}
\usepackage{diagbox}
\usepackage{verbatim}
\usepackage{makecell}

\title{SUMBot: Summarizing Context in Open-Domain Dialogue Systems}
\name{Rui Ribeiro, Luísa Coheur}
\address{
  INESC-ID, Lisboa, Portugal\\
  Instituto Superior Técnico, Universidade de Lisboa, Portugal}
\email{rui.m.ribeiro@tecnico.ulisboa.pt, luisa.coheur@tecnico.ulisboa.pt}

\begin{document}

\maketitle
\begin{abstract}
In this paper, we investigate the problem of including relevant information as context in open-domain dialogue systems. 
Most models struggle to identify and incorporate important knowledge from dialogues and simply use the entire turns as context, which increases the size of the input fed to the model with unnecessary information.
Additionally, due to the input size limitation of a few hundred tokens of large pre-trained models, regions of the history are not included and informative parts from the dialogue may be omitted.
In order to surpass this problem, we introduce a simple method that substitutes part of the context with a summary instead of the whole history, which increases the ability of models to keep track of all the previous relevant information.
We show that the inclusion of a summary may improve the answer generation task and discuss some examples to further understand the system's weaknesses.
\end{abstract}

\section{Introduction}
\label{section:introduction}

Chit-chat systems have become more and more prominent with the emergence of large pre-trained models and the increased access to public libraries \cite{HUGGINGFACE, ALLENLP, PARLAI} that allow to easily train and deploy these models.
Specifically, new advances have shown promising progress in the dialogue generation task, as these systems became more competent at providing human-like answers.
However, these deep-learning systems tend to generate generic responses which are repetitive or incoherent with the context, particularly when conversations attain many interactions and contain long turns.

Recent approaches have studied the ability of deep generative models to capture relevant information from the dialogue context \cite{EMPIRICALCONTEXT, CONTEXTAWARE}. 
They have found that these models do not efficiently make use of all parts from the dialogue history and tend to ignore relevant turn information.
Other approaches \cite{PRETRAINING, CONTEXTDA, TEMPLATE, CONVERT} have attempted to represent the context and leverage the resulting representations to various dialogue tasks.
However, none of these approaches has studied the substitution of the context with a summary.

\begin{figure}[t]
\centering
\includegraphics[width=0.5\textwidth]{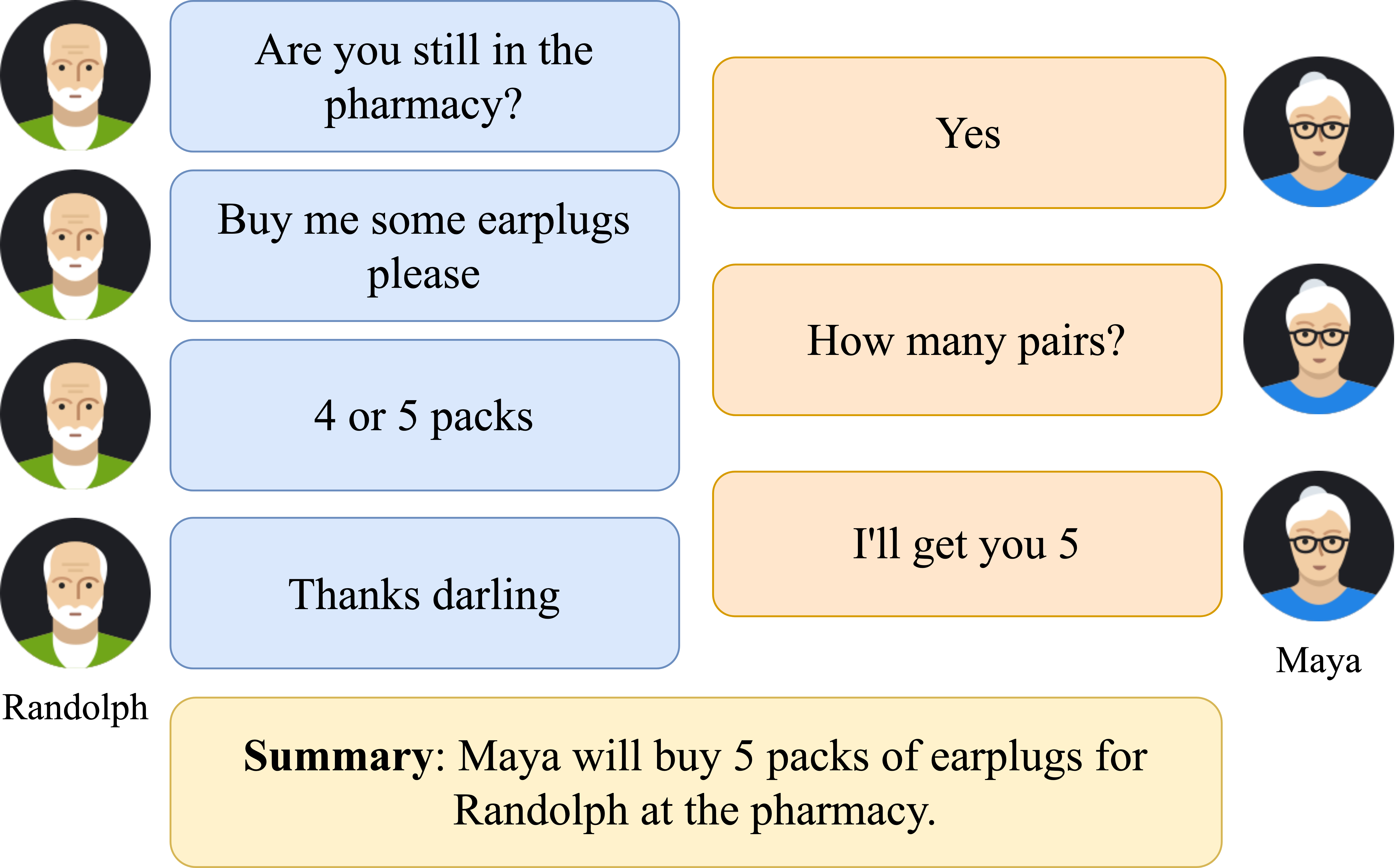}
\caption{Example of a dialogue between two speakers and the respective summary on the SAMSum dataset.}
\label{fig:samsum}
\end{figure}

In this paper, we investigate the importance of encapsulating complete dialogue utterances into a summary and reducing the context size in the open-domain dialogue task. 
We attempt to answer the following question: can a summary of the previous context include all the important information and also decrease the input size fed to a model?
To answer this question, we propose a simple yet effective method that incorporates summaries of the previous turns that are not included as input.
More specifically, apart from the user request, we only include a few complete speaker turns, and the remaining turns are compiled into a summary that describes succinctly the omitted utterances.
We train different versions of the model where we change the number of complete utterances provided, which may vary between 0 and 10.
This procedure allows us to analyze if the inclusion of summaries is an effective strategy and if the summaries become a valuable choice as substitutes for the complete turns. 

\begin{figure*}[t]
  \includegraphics[width=\textwidth]{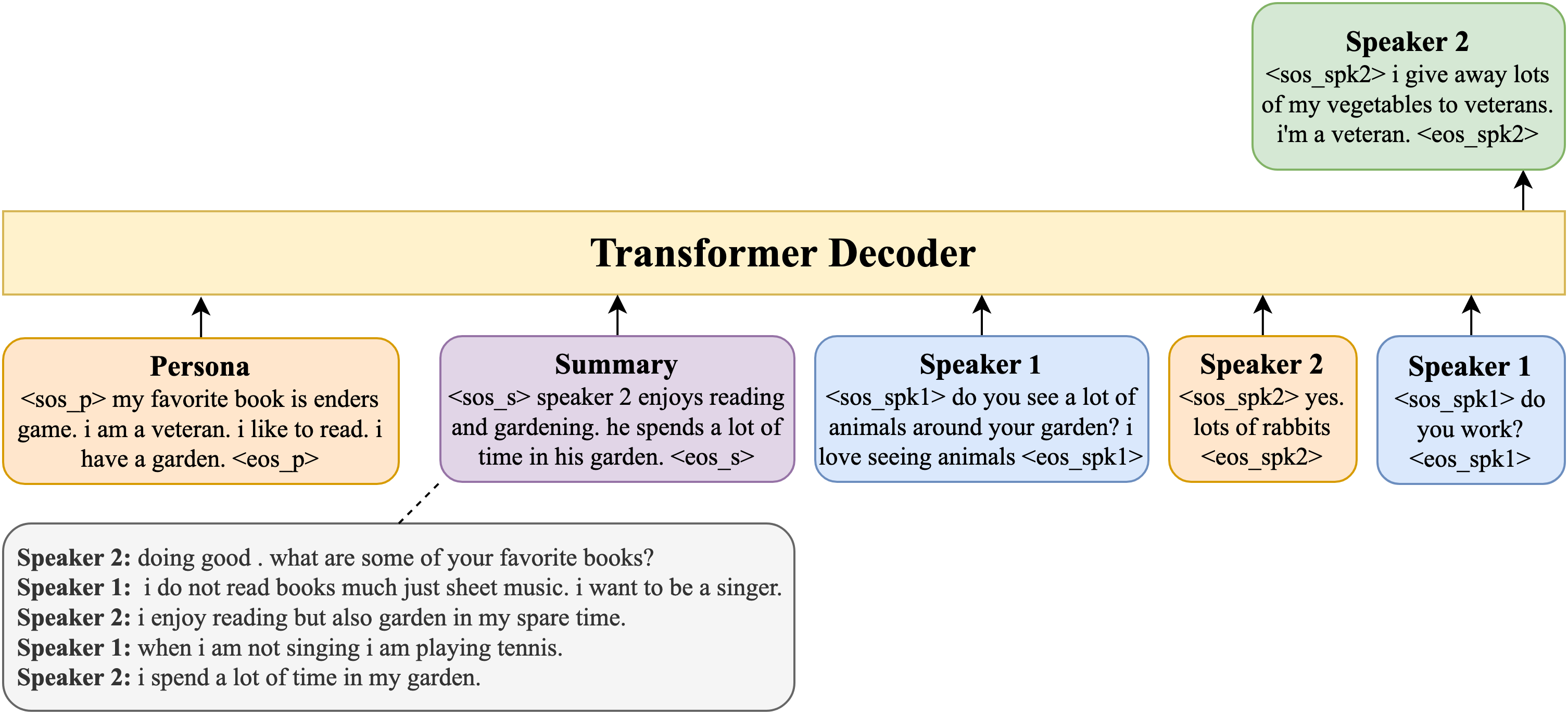}
  \caption{Example of an input fed to our model and the corresponding generated answer. Here, the summary represents the whole history that was not included as full turns.}
    \label{fig:decoder}
\end{figure*}

The training is divided into two independent stages: first, we fine-tune BART \cite{BART} in the SAMSum corpus \cite{SAMSUM} and use it to generate summaries for the dialogue context.
Figure \ref{fig:samsum} shows an example of a dialogue from this dataset.
Then, we fine-tune DialoGPT decoder \cite{DIALOGPT} with the summaries from the previous stage by incorporating them with the dialogue between both speakers.

We evaluate our model on the open-domain Persona-Chat dataset \cite{PERSONA} and observe that the inclusion of the summaries may improve the overall results. 
We also analyze if the summaries are proper substitutes for the dialogue history and discuss possible flaws that can decrease the performance of the generation model.


\section{Related Work}

Since the introduction of encoder-decoder models \cite{SEQ2SEQ, MT-SEQ2SEQ},
chit-chat dialogue systems have been in constant evolution and are more capable of generating fluent and human-like sentences.
In these systems, the encoder extracts important features from the utterances and passes that information to a decoder that generates a response.

Considering that our approach attempts to provide a proper substitute for the dialogue history, the related work that becomes more relevant focuses on studying and representing the context in the dialogue task.
\cite{EMPIRICALCONTEXT} study the aptitude of encoder-decoder models based on RNNs and Transformers to interpret and understand the dialogue context.
The authors introduce synthetic perturbations to the history such as word shuffling and sentence reordering and discover that most models are insensitive to these input perturbations, which suggests that they fail to capture the dialogue dynamics between turns.

Few works attempt to identify relevant information in conversation context. \cite{CONVERT} introduce ConveRT, a lightweight framework to encode multi-turn context where it is possible to transfer learned representations to other dialogue tasks.
They apply their model to the response selection task and achieve strong results in three response retrieval datasets.
\cite{CONTEXTMATTERS} apply ConveRT to encode the dialogue context and merge the resultant representation with the user request to generate appropriate and context-aware responses.
They show that using a better context representation increases the ability to generate more diverse responses, and, as we also discuss in this paper, a longer context does not always correlate to an increase in the quality of responses. 

These approaches have introduced different techniques to represent and embody context into dialogue models. 
In this paper, we will also attempt to represent dialogue but use generated summaries to substitute part of the dialogue history.

\section{Method}

\subsection{Summary Generation}

\begin{table*}[t]
\centering
\begin{tabular}{lllllllll}
\toprule
\thead{\textbf{Complete} \\ \textbf{Turns}} & \thead{\textbf{Includes} \\ \textbf{Summary?}} & \thead{\textbf{BLEU-4 (\%)}} & \thead{\textbf{ROUGE-1 (\%)}} & \thead{\textbf{ROUGE-2 (\%)}} & \thead{\textbf{ROUGE-L (\%)}} & \thead{\textbf{Avg.} \\ \textbf{Length}} & \thead{\textbf{Max.} \\ \textbf{Length}}\\ \midrule
0 & No & 3.70 & 18.4 & 4.23 & 17.6 & 71 & 115 \\ %
2 & No & 3.94 & 19.2 & 4.55 & 18.3 & 96 & 291 \\ %
4 & No & 3.86 & 19.4 & 4.62 & 18.5 & 118 & 309 \\ %
6 & No & 4.03 & 19.6 & 4.30 & 18.6 & 136 & 366 \\ %
8 & No & 3.32 & 19.3 & 4.50 & 18.4 & 150 & 274 \\ %
10 & No & 3.89 & 18.0 & 3.66 & 17.2 & 160 & 434 \\ \midrule %
0 & Yes & 3.76 & 18.7 & 4.23 & 17.9 & 86  & 115 \\ %
2 & Yes & 3.95 & 19.5 & 4.72 & 18.5 & 107 & 305 \\ %
4 & Yes & 3.95 & 19.1 & 4.19 & 18.2 & 127 & 349 \\ %
6 & Yes & 3.73 & 18.9 & 4.28 & 18.0 & 140 & 376 \\ %
8 & Yes & 4.11 & 19.5 & 4.44 & 18.6 & 153 & 380 \\ %
10 & Yes & 4.05 & 19.3 & 4.13 & 18.3 & 162 & 386 \\ 
\bottomrule %
\end{tabular}
\vspace*{2mm}
\caption{
Results for BLEU and ROUGE metrics on the Persona-Chat dataset.
We also report the average and maximum lengths for the contexts in terms of tokens.}
\label{table:main}
\end{table*}

As discussed earlier, our goal is to substitute parts of the dialogue history with a summary that could encompass all the relevant information.
To summarize the dialogue context, we use BART-large \cite{BART}, a transformer architecture with a bidirectional encoder similar to BERT \cite{BERT} and a decoder similar to GPT \cite{GPT2}.
This model is pre-trained on the English Wikipedia and BookCorpus datasets.
Since the goal of this stage is to summarize parts of dialogues, we fine-tune the model on the SAMSum corpus \cite{SAMSUM}, an abstractive summarization dataset where the task is to summarize a dialogue between two interlocutors.
After that, we use the resultant model to generate the summaries for every turn of the dataset.

In the Persona-Chat dataset, each dialogue is sustained with persona information from Speaker 2, the answering bot, and the task becomes delivering an appropriate and consistent answer given the dialogue context and the persona.
Thus, instead of summarizing the turns from both speakers, we summarize only the turns from Speaker 2, which, intuitively, correspond to the persona aspects that were mentioned earlier in the dialogue context. 


\begin{figure}[h]
    \centering
    \includegraphics[width=0.47\textwidth]{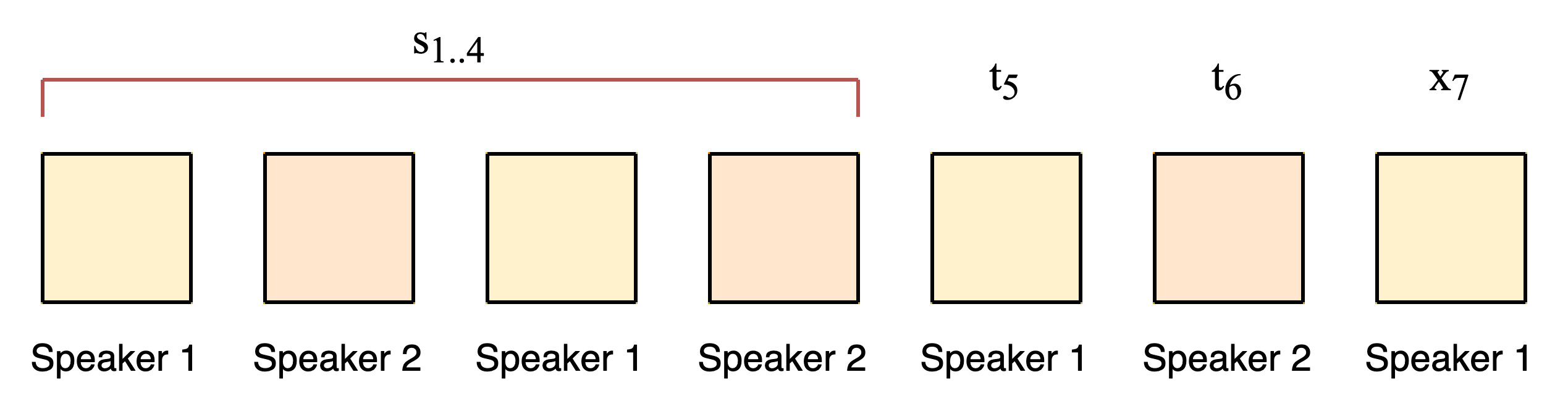}
    \caption{An example of the input for the decoder at step 7. For instance, this setup corresponds to the row of 2 complete turns with a summary from Table \ref{table:main}, where the first 4 sentences are summarized and only the previous 2 turns are included.}
    \label{fig:input}
\end{figure}

\subsection{Decoder Fine-Tuning}
\label{sec:finetune}

In this stage, we fine-tune a GPT-2 transformer decoder in the Persona-Chat dataset.
An example with a real dialogue from the dataset can be seen in Figure \ref{fig:decoder}, where the last 3 sentences are included as full utterances, while the remaining dialogue is abbreviated into the summary.
We use a pre-trained version trained on large corpora of dialogues, DialoGPT \cite{DIALOGPT}, which produces more relevant and context-consistent answers in comparison to the original pre-trained version.

The input is structured as follows: consider a dialogue \(d\) with \(n\) turns and a persona \(p\).
Then, the input \(y_n\) at the \(n\)-th turn can be described as:

\[ y_n = \{p, s_{1..n-i-1}, t_{n-i}, ..., t_{n-1}, x_{n}\},\]

where \(i\) is the number of complete turns, \(s_{1..n-i-1}\) is the summary of the turns omitted, \(t_{n-i}, ..., t_{n-1}\) correspond to the complete turns included, and \(x\) is the request from Speaker 1.
This structure can be observed in Figure \ref{fig:input}.
The model then generates an appropriate response, \(r_t\), for the \(n\)-th turn of Speaker 2 according to the distribution \(p(r_n | t_n)\).

With this, our model can both reduce the input size and keep track of the context long ago.
We create special tokens for each part of the input in order to help the model distinguish between the different segments.

\section{Experiments}

\subsection{Experimental Setup}

In order to compare and evaluate the impact of adding summaries as input, we train different versions of the model where we vary the number of complete turns that are provided. 
As discussed in Section \ref{sec:finetune}, we use DialoGPT, a version of GPT-2 trained specially for the dialogue generation task, which contains 12 layers of decoder Transformer blocks.
The maximum input size is 1024 and we generate an answer with a maximum size of 200.
We use Adam as the optimizer with a learning rate of \(6.25e^{-5}\), and train the model for 5 epochs with patience 1.
We use HuggingFace's library \cite{HUGGINGFACE} which provides an implementation with a language modeling head on top of the GPT-2 decoder, and generate the answer using a greedy search approach, where the next word selected is the one with the highest probability.

As mentioned earlier, the dataset used for the generation task is the Persona-Chat \cite{PERSONA}, which contains 1155 possible personas consisting of at least 5 profile sentences, such as \textit{``I like to go hunting''} or \textit{``my favorite holiday is Halloween''}.
We report BLEU \cite{BLEU} and ROUGE \cite{ROUGE}, both automatic metrics that measure fluency by comparing the word occurrences between the generated and the ground truth responses.

\begin{table*}[t]
\small
\centering
\begin{tabular}{lllllllll}
\toprule
        & & \multicolumn{7}{c}{\textbf{Context Size}} \\ \cmidrule{3-9}
\thead{\textbf{Complete} \\ \textbf{Turns}} & \thead{\textbf{Includes} \\ \textbf{Summary?}} & 0 & 2 & 4 & 6 & 8 & 10 & 12 \\ \midrule
0 & No & {7.20} & 4.63          & {4.04} & 3.73          & 2.76          & 2.39          & 2.31          \\ 
0 & Yes & 6.33          & {4.71} & 3.65          & {4.18} & {3.06} & {2.46} & {2.44} \\ \midrule
2 & No & - & {5.09} & {4.01} & 3.86          & 2.90          & {2.80} & 2.61          \\ 
2 & Yes & - & 4.77 & 3.87          & {4.05} & {3.12} & 2.79          & {2.99} \\ \midrule
4 & No & - & -& {4.99} & {4.21} & {4.07} & 3.25          & 2.75          \\
4 & Yes & - & - & 4.87          & 3.77          & 3.99          & {3.45} & {2.86} \\ \midrule
6 & No & - & - & - & {4.60} & {3.39} & {3.16} & {2.97} \\
6 & Yes & - & - & - & 3.66          & 2.94          & 2.85          & 2.37          \\ \midrule
8 & No & - & - & - & - & 2.67          & {2.88} & 1.95          \\
8 & Yes & - & - & - & - & {3.51} & 2.78          & {3.06} \\ \midrule
10 & No & - & - & - & - & - & 2.93   & {2.64} \\ 
10 & Yes & - & - & - & - & - & {2.97} & 2.57          \\ \bottomrule
\end{tabular}
\vspace*{2mm}
\caption{BLEU score per context size.
We reorganize the individual results for different context sizes.
The results show that when the number of complete turns is low (for instance, row 0 and 2), including a summary improves the results, especially when the context size is longer.
We omit the results where there is no possible summary, as all context is included as complete turns.}
\label{table:turns}
\end{table*}

\begin{table}[!b]
\small
\centering
\begin{tabular}{p{0.4\textwidth}}
 \textbf{Dialogue}\\
 1. {[SPEAKER 1]} Hi, how are you doing? I'm getting ready to do some cheetah chasing to stay in shape.\\
 2. {[SPEAKER 2]} You must be very fast. Hunting is one of my favorite hobbies.\\
 3. {[SPEAKER 1]} I am! For my hobby I like to do canning or some whittling. \\
 4. {[SPEAKER 2]} I also remodel homes when i am not out bow hunting.\\
 5. {[SPEAKER 1]} That's neat. When I was in high school I placed 6th in 100m dash !\\
 6. {[SPEAKER 2]} I also remodel homes when i am not out bow hunting.\\
 7. {[SPEAKER 1]} That's neat. When I was in high school I placed 6th in 100m dash !\\
 \midrule
 \textbf{Summary (1-3)}\\
 Speaker 2 likes hunting, it's one of his favourite hobbies.\\
 \midrule
 \textbf{Summary (1-5)}\\
 Speaker 2 enjoys hunting and remodeling when he's not doing it.\\
 \midrule
 \textbf{Summary (1-7)}\\
 Speaker 2's favorite hobby is hunting. He also likes to remodel when he's not hunting.\\

\end{tabular}
\vspace{2mm}
\caption{Example of generated summaries from our model.}
\label{table:dialogue}
\end{table}

\subsection{Results and Discussion}

In Table \ref{table:main}, we present the results of the experiments, where the first half contains the versions that do not include a summary and the second part contains the versions with a summary.
We additionally include the average and maximum tokens per utterance for each version.
The complete turns increase with step 2 due to the request-response nature of dialogue.

We observe that providing more context to the model does not directly reflect on higher results. 
For instance, we observe a decrease in score at rows 4 and 8 of complete turns without summary.
The results show that, in general, the inclusion of the summary does improve a few points in BLEU except in the row 6 of complete turns, where it achieves a lower score.
This demonstrates that, in some cases, it is not certain that the summaries improve the results.

In Table \ref{table:turns}, we rearrange the results per context length, i.e., the stage of the dialogue at the current step.
For instance, at context size 2, we consider that the bot is generating the 3rd utterance of the dialogue.




\subsubsection{Quality of Summaries} 
The results show that, in general, the inclusion of a summary improves the generation results when the number of complete turns are the same.
However, if we consider the summary as a substitute of the complete turns, it is not completely certain that the model is able to encapsulate all the information.
For instance, we observe that using only 2 complete turns with a summary outperforms all except using 6 complete turns without a summary.
However, with the decrease in performance at 6 complete turns with summary, it reaches the results from using no summary neither complete turns.

We believe that this happens due to the dependency of the dialogue generation model on the quality of the generated summaries.
As mentioned earlier, we use a pre-trained model to produce summaries for the dialogue context, which implies that the errors from the summaries are propagated to the decoder, which may contribute to a performance decrease.
In Table \ref{table:dialogue}, we provide some examples of generated summaries for a dialogue where we can observe this issue: although it is evident that the first and the last summaries successfully encapsulate the information, the second summary is incoherent.

In other scenarios, the summary focuses on irrelevant information such as greetings \textit{``Speaker 2 wants to know how are you doing''} or excluded from the summaries significant information.
By omitting this information, the model is not able to understand that this knowledge was already mentioned in a previous turn, which leads to the generation of similar and repetitive responses, and may lead to a worse generation quality.

\subsubsection{Input Length} 
We report the average and maximum size of the input fed to the model.
We observe that the maximum length of the inputs does not come closer to the maximum input size of 1024 from GPT-2.
However, if we are required to reduce the model's maximum input size or if we consider a scenario where the dialogues were very extensive, the inclusion of the summary would allow us to reduce the size and include information from all the previous context.

\subsubsection{Impact at Longer Contexts} 
Finally, we performed an extensive analysis of the results by calculating the overall score for each context size.
The results in Table \ref{table:turns} show that when the number of complete turns provided is small and the context size grows, the models with summary outperform the models without summary.
This follows our intuition that the summaries are able to encapsulate the part of the dialogue that is not included as input.
However, this is not always true: at 6 complete turns, the model without summary outperforms in almost all the different context sizes.
As already discussed, we believe that this happens due to the reliance on the quality of the summaries and to the uncertain nature of the dialogue generation model.

\section{Conclusion and Future Work}

In this paper, we present a simple yet effective method for representing dialogue context in the open-domain setting.
We show that, in some cases, it is possible to improve dialogue generation and reduce the size of the input, however, due to the poor quality of the summaries and the generation model, the model may achieve a lower score when compared to the baseline without summary.
In future work, we would like to study other approaches to better summarize the relevant information from the context and extend the use of dialogue summarization to the task-oriented setting.

\section{Acknowledgements}

Work supported by FCT (project UIDB/50021/2020), ANI and CMU Portugal (project 045909 - MAIA), and the European Regional Development Fund (LISBOA-01-0247-FEDER-045385)

\bibliographystyle{IEEEtran}

\bibliography{mybib}


\end{document}